\documentclass{article} 
\usepackage{iclr2020_conference,times}

\usepackage{amsmath,amsfonts,bm}

\def\eqref#1{equation~\ref{#1}}

\def\1{\bm{1}}

\DeclareMathAlphabet{\mathsfit}{\encodingdefault}{\sfdefault}{m}{sl}
\SetMathAlphabet{\mathsfit}{bold}{\encodingdefault}{\sfdefault}{bx}{n}

\usepackage{hyperref}
\usepackage{url}

\title{A Hierarchical Transformer for Unsupervised Parsing }

\author{Ashok Thillaisundaram\\
        BenevolentAI\\
	    4-8 Maple Street\\
	    London W1T 5HD\\
	    {\tt ashok@benevolent.ai}
}

\iclrfinalcopy 
\begin{document}

\maketitle

\begin{abstract}
The underlying structure of natural language is hierarchical; words combine into phrases,  which in turn form clauses. An awareness of this hierarchical structure can aid machine learning  models in  performing  many  linguistic tasks. However, most such models just process text sequentially and there is no bias towards learning hierarchical structure encoded into their architecture. In this  paper, we extend the recent transformer model (Vaswani  et  al., 2017) by enabling it to learn hierarchical representations. To achieve this, we adapt the ordering mechanism introduced in Shen et al., 2018, to the self-attention module of the transformer architecture. We train our new model on language modelling and then apply it to the task of unsupervised parsing. We achieve reasonable results on the freely available subset of the WSJ10 dataset with an F1-score of about 50\%.
\end{abstract}

\section{Introduction and Summary of Main Results}

Natural language may be spoken and written as a sequence of words but the underlying syntactic structure in language is hierarchical. If a sentence is decomposed into its constituents, the resulting structure is tree-like; we refer to this as the sentence's constituency-based parse tree. These parse trees are crucial for many tasks in natural language processing. Obtaining the parse trees for a large corpus of text can be done via human annotation; however, this can be very time-consuming. Alternatively, supervised syntactic parsers can be used. Training these parsers will still require large amounts of annotated text and, for many languages, such datasets do not exist. Unsupervised parsing is therefore of much interest especially in languages that are not as extensively studied; training these unsupervised models would not require annotated parse trees.

One method for training unsupervised parsers is inspired by human language acquisition. When native speakers learn their mother tongue, they usually gain a rough understanding of the underlying sentence structure even though they are not explicitly given syntactic information. This naturally motivates the question of whether the same can be done with neural language models: Can deep learning models learn the latent hierarchical structure of a sentence while being trained for next word prediction? This would involve modifying the architecture of a deep learning model to incorporate an inductive bias towards learning hierarchical representations. These trained hierarchical neural language models could then be used for unsupervised parsing. 

In our work, we modify the well-known transformer architecture \citep{Vaswani17} to incorporate a bias towards learning latent hierarchical representations. We train our modified unidirectional transformer encoder on language modelling and then apply it to the task of unsupervised parsing. We achieve reasonable results for unsupervised parsing on the freely available subset of the WSJ10 dataset with an F1-score of about 50\%. 

Our modification of the transformer architecture is an adaptation of the idea introduced in \cite{Shen18}; here, the authors presented a model called the Ordered Neurons Long Short Term Memory (ON-LSTM) network. For the standard LSTM, as each word is processed, information is written to and deleted from the cell state as dictated by the input and forget gates. For the ON-LSTM, the neurons in the cell state are ordered in the sense that if a neuron at a higher position is deleted, all preceding neurons must also be deleted. The analogy here is that if a larger constituent in a sentence ends (e.g. a clause), all smaller constituents contained within it (e.g. the phrases within the clause) must also end. This mechanism (along with a similar mechanism for writing to the cell state) ensures that information relating to smaller constituents on shorter timescales is written to neurons at lower positions, while information regarding larger constituents which is relevant over longer timescales is written to neurons at higher positions. From this ordering of neurons, it is possible to reconstruct the parse tree of the sentence by noting all the various points when a constituent is closed. 

Now, for the transformer architecture, one of its notable innovations is that it no longer requires recurrence. Given this, the idea from the ON-LSTM where you process each word sequentially, writing to and deleting from the cell state as described above, cannot be adapted in a straightforward manner to the transformer architecture. Instead, we modify the self-attention module in the transformer to achieve a similar effect. For standard self-attention, a given word attends to all the previous words, then the representations for all the words are averaged according to the attention weights. For our modified self-attention, the representation for each word is also multiplied by an input gate then all word representations are summed weighted by the attention weights. A forget gate is then applied on top of this deleting all information up till a certain neuron (analogous to closing a large constituent if one is present). The end result here is that the model is biased towards writing information that is crucial over longer timescales to neurons at higher positions similar to the ON-LSTM model. It performs comparably to the ON-LSTM on the same subset of the WSJ10 dataset. Importantly, we chose to modify the unidirectional transformer instead of the bidirectional transformer because the former more closely mimics the behaviour of an LSTM; it processes all previous words before outputting a prediction for the current position. It may be possible to adapt the ordering mechanism from the ON-LSTM to the bidirectional transformer but this seems less straightforward.

To place our work in context, we briefly mention some related work. There have been various other models proposed in the last few years which include an inductive bias towards learning hierarchical structures. Both \cite{Koutnik14} and \cite{Chung16} proposed other methods which allow RNNs to learn latent hierarchies. In terms of variations of the transformer architecture, recent work by \cite{Wang19} introduced the Tree Transformer which provides another method of enabling a transformer to learn tree-like structures. However, in their case, the depth of the tree must be specified in advance while our model can learn trees of arbitrary depth.

\section*{Acknowledgements}
 We would like to thank Angus Brayne, Julien Fauqueur, Juha Iso-Sipila, Thomas Edlich, Benedek Fabian, and Marwin Segler for proofreading and/or engineering support.

\bibliography{iclr2020_conference}
\bibliographystyle{iclr2020_conference}

\end{document}